\def\year{2021}\relax
\documentclass[letterpaper]{article} 
\usepackage{aaai21}  
\usepackage{times}  
\usepackage{helvet} 
\usepackage{courier}  
\usepackage[hyphens]{url}  
\usepackage{graphicx} 
\usepackage{natbib}  
\usepackage{caption} 
\usepackage{tabulary}
\usepackage{booktabs}
\usepackage[switch]{lineno}
\input{def.set}

\title{SeCo: Exploring Sequence Supervision for\\ Unsupervised Representation Learning}
\author{Ting Yao, Yiheng Zhang, Zhaofan Qiu, Yingwei Pan, Tao Mei\\}
\affiliations{JD AI Research, Beijing, China\\
\{tingyao.ustc, yihengzhang.chn, zhaofanqiu, panyw.ustc\}@gmail.com, tmei@jd.com}

\begin{document}

\maketitle
\begin{abstract}
A steady momentum of innovations and breakthroughs has convincingly pushed the limits of unsupervised image representation learning. Compared to static 2D images, video has one more dimension (time). The inherent supervision existing in such sequential structure offers a fertile ground for building unsupervised learning models. In this paper, we compose a trilogy of exploring the basic and generic supervision in the sequence from spatial, spatiotemporal and sequential perspectives. We materialize the supervisory signals through determining whether a pair of samples is from one frame or from one video, and whether a triplet of samples is in the correct temporal order. We uniquely regard the signals as the foundation in contrastive learning and derive a particular form named Sequence Contrastive Learning (SeCo). SeCo shows superior results under the linear protocol on action~recognition (Kinetics), untrimmed activity recognition (ActivityNet) and object tracking (OTB-100). More remarkably, SeCo demonstrates considerable improvements over recent unsupervised pre-training techniques, and leads the accuracy by 2.96\% and 6.47\% against fully-supervised ImageNet pre-training in action recognition task on UCF101 and HMDB51, respectively. Source code is available at \url{https://github.com/YihengZhang-CV/SeCo-Sequence-Contrastive-Learning}.
\end{abstract}

\section{Introduction}

Supervised learning has made significant progress and is still dominant in visual representation learning. Despite having high quantitative performances, the achievements rely heavily on the requirement to have large number of expert annotations for training deep neural networks, and the acquisition of annotations is an intellectually expensive and time-consuming process. Moreover, the representations especially learnt on very specific tasks in a supervised manner may suffer from generalization problem and transfer poorly to other objectives. In contrast, unsupervised representation learning alleviates the issues by completely exploiting the inherent structures and correlations from the data as the supervision. This is particularly applicable to video, which is an information-intensive media with spatiotemporal coherence and variation. Such facts motivate the explorations of building unsupervised learning models to yield powerful and generic representations.

\begin{figure*}[!tb]
\centering {\includegraphics[width=0.92\textwidth]{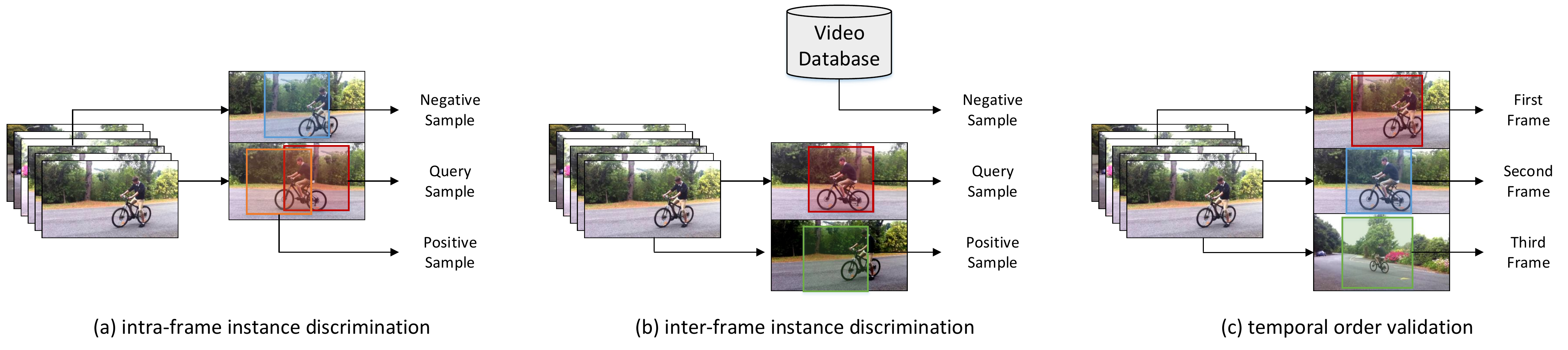}}
\caption{One proxy task on each type of supervision in video sequence for unsupervised learning.}
\label{fig:intro}
\end{figure*}

The supervision in the video sequence generally originates from three types: spatial, spatiotemporal, and sequential. In between, spatial supervision is derived from the structures in static frame, spatiotemporal supervision reflects the correlation across different frames, and sequential supervision verifies the temporal coherence. In the literature, unsupervised learning methods for videos often involve different proxy tasks, e.g., predicting the pixel-level displacement across consecutive frames \cite{Liu2017VideoFS,Vondrick2016GeneratingVW,Wang2019SelfSupervisedSR}, or reconstructing/predicting the input/future frame through decoder \cite{Han2019VideoRL,Luo2017UnsupervisedLO,Srivastava2015UnsupervisedLO}, and execute representation learning through optimizing such tasks with the supervision. Here, without loss of simplicity and generality, we present one simple proxy task on each type of supervision. From the spatial standpoint, we extend the instance discrimination task in \cite{he2019momentum,wu2018unsupervised,cai2020joint} to an intra-frame instance discrimination task, which distinguishes whether two frame patches are from the same video frame, as shown in Figure \ref{fig:intro}(a). From the spatiotemporal perspective, we remould an inter-frame instance discrimination task, which determines whether two frame patches are derived from an identical video, as depicted in Figure \ref{fig:intro}(b). For sequential supervision, we develop a task of temporal order validation (Figure \ref{fig:intro}(c)) and verify whether a series of frame patches are in the correct temporal order.

To materialize the exploitation of supervision in the sequence through the three proxy tasks, we present a new Sequence Contrastive Learning (SeCo) approach for unsupervised representation learning. Considering that contrastive learning is at the core of recent advances \cite{he2019momentum,wu2018unsupervised} on unsupervised learning, we build SeCo on this recipe. The basic principle is to make positive/negative query-key pairs similar/dissimilar. Specifically, for each video, we randomly sample three frames and take either first frame or the last frame in time order as the ``anchor'' frame. In both intra-frame and inter-frame instance discrimination tasks, we perform data augmentation on the ``anchor'' frame to generate two image patches. One is taken as query and the other patch plus the augmentations of another two frames are used as keys. Moreover, inspired by \cite{he2019momentum}, we additionally build a memory to track keys across mini-batches for inter-frame instance discrimination task. InfoNCE \cite{oord2018representation}, as one form of contrastive formulation, serves as the loss function in the two tasks. For the task of temporal order validation, we take the ``anchor'' frame as query and the rest as keys. We involve a linear classifier to predict if the query is in front of or behind keys (two-class classification). The classifier takes the concatenation of the features of query and keys as the input and is learnt via cross-entropy loss. Overall,~SeCo is end-to-end trained by jointly optimizing the three proxy~tasks.

The main contribution of this work is the proposal of exploring sequence supervision for unsupervised representation learning. Ours is among the first to systematically analyze the supervisory signals behind the rich structures in video sequence. This also leads to the elegant views of how to design simple proxy tasks which perform as a prism through which to leverage the supervision, and how to nicely capitalize on such proxy tasks for learning a generic representation, which are problems not yet fully understood. We demonstrate the effectiveness of SeCo on several downstream video applications and SeCo unsupervised pre-training also surpasses the ImageNet supervised~pre-training on two video benchmarks for action recognition.

\section{Related Work}
\textbf{Unsupervised Learning from Video} aims to learn a generic representation without using any explicit semantic labels, which can be briefly grouped into three major categories.
The first group learns feature representation by leveraging appearance variations in videos. For example, the most common constraint is to enforce the learnt representation to be temporally smooth \cite{mobahi2009deep,pan2016learning,wang2015unsupervised,zou2012deep}. Moving beyond only temporal smoothness, ego-motion constraints \cite{agrawal2015learning,jayaraman2015learning}, object tracking \cite{wang2015unsupervised} and temporal order verification \cite{misra2016shuffle} have been employed to further regularize the learning process. The recent works also attempt to learn the representation by predicting the pixel-level displacement across consecutive frames \cite{Liu2017VideoFS,Vondrick2016GeneratingVW,Wang2019SelfSupervisedSR}.
The second group focuses on temporal prediction and frame reconstruction tasks \cite{finn2016unsupervised,Han2019VideoRL,Luo2017UnsupervisedLO,Srivastava2015UnsupervisedLO}. \cite{Srivastava2015UnsupervisedLO} utilizes a LSTM-based encoder-decoder structure to reconstruct current frame or predict future frames. \cite{finn2016unsupervised} further upgrades \cite{Srivastava2015UnsupervisedLO} by merging appearance information from previous frames with motion cues. Luo \textit{et al.} \cite{Luo2017UnsupervisedLO} present to describe the motion between frames as a sequence of atomic 3D flows to predict long-term motion. More recently, \cite{Han2019VideoRL} learns a dense encoding of spatio-temporal blocks by recurrently predicting future representations.
The third group attempts to predict the transformation parameters from the transformed video \cite{Ahsan2019VideoJU,Jing2018SelfSupervisedSF}. Jing \textit{et al.} \cite{Jing2018SelfSupervisedSF} introduce a pretext task which is defined as the prediction of the rotations applied to videos. Ahsan \textit{et al.} divide multiple video frames into grids of patches and train a network to solve jigsaw puzzles on these patches from multiple frames in \cite{Ahsan2019VideoJU}.

\textbf{Self-Supervised Learning} is a form of unsupervised learning. It relies only on the data itself for some form of supervision without human-annotated labels. One mainstream of self-supervised learning focuses on the pretext tasks which are designed under various scenarios only for learning a good data representation. Some pretext tasks, e.g., relative patch prediction \cite{doersch2015unsupervised,goyal2019scaling,noroozi2016unsupervised}, affine transformation prediction \cite{gidaris2018unsupervised}, and colorization \cite{deshpande2015learning,zhang2016colorful}, are proven to be helpful for representation learning. Recently, contrastive learning is at the core of several works on self-supervised learning \cite{bachman2019learning,hjelm2019learning,wu2018unsupervised}. The design principle is to maximize/minimize the similarity between the instances in positive/negative pairs and various pretext tasks can be represented in a contrastive manner. For instance, both contrastive multiview coding (CMC) \cite{tian2019contrastive} and colorization \cite{deshpande2015learning} attempt to make the representation be invariant to the color in images. For self-supervised contrastive video representation learning, Contrastive Predictive Coding (CPC) \cite{lorre2020temporal} is proposed to learn long-term relations underlying the raw signal and predict the latent representation of future segments in the video.
The most closely related work is Momentum Contrast (MoCo) \cite{he2019momentum}, which builds dynamic dictionaries for contrastive learning and leverage the instance discrimination task for unsupervised image feature learning.
Our method is different in the way that we explore the generic supervision in the video sequence from spatial, spatiotemporal, and sequential perspectives, for unsupervised video representation learning.

\section{Preliminary---Contrastive Learning for\\ Unsupervised Feature Learning}
We briefly review contrastive learning and its recent practical variant (MoCo \cite{he2019momentum}), which learn feature embedding in an unsupervised manner by making positive/negative query-key pairs similar/dissimilar. Formally, suppose we have an encoded {\em query} $\bq \in {\mathcal R}^d$, and a group of encoded {\em key} vectors $\mathcal{K}=\{\bk^{+},\bk_{1}^{-},\bk_{2}^{-},...,\bk_{K}^{-}\}$ consisting of one positive key $\bk^{+}\in {\mathcal R}^d$ and $K$ negative keys $\mathcal{K}^-=\{\bk_{i}^{-}\}$, where $d$ denotes the dimension of the embedding space. Note that the positive key $\bk_i^{+}$ comes from the same distribution as the query $\bq$, while the negative keys are derived from an alternative noise distribution. The objective of typical contrastive loss is to reflect the incompatibility of each query-key pair: returns low value when query $\bq$ is similar to its positive key $\bk^{+}$ and remains distinct to all negative keys $\{\bk_{i}^{-}\}$. By measuring the query-key similarity via~dot product, a prevailing form of contrastive loss (InfoNCE \cite{oord2018representation}) is calculated in a softmax formulation:
\begin{equation}\scriptsize
\label{eq:cpc}
\mathcal{L}_{NCE}(\bq,\bk^{+},\mathcal{K}^-)=-\log \frac{\exp(\bq^T \bk^{+}/\tau)}  {\exp(\bq^T \bk^{+}/\tau)+\sum_{i=1}^{K} \exp(\bq^T \bk_{i}^{-}/\tau)},
\end{equation}
where $\tau$ is the temperature hyper-parameter. The rationale behind such formulation is to train a classifier that could correctly classify query $\bq$ as positive key $\bk^{+}$.

Because no human-annotated labels are available in unsupervised setting, one common practice is to produce two different augmentations ($x_q$, $x^+_k$) from the same instance (an image $x$), which correspond to the query $\bq$ and positive key $\bk^{+}$. The augmentations of other instances/images $\{x_{k}^{-}\}$ are taken as the negative keys $\{\bk_{i}^{-}\}$.
In this way, a simple instance discrimination task is designed for unsupervised visual representation learning: determining whether two image patches are derived from the same image.
In the implementation, two encoders (query encoder $f_q$ and key encoder $f_k$) are utilized to map query image $x_q$ and each positive/negative key image $x_k$ into the embedding space (i.e., $\bq=f_q(x_q)$, $\bk=f_k(x_k)$). Recently, MoCo \cite{he2019momentum} strengthens contrastive learning by involving an extreme large number of negative keys via maintaining a dynamic memory to track the keys across mini-batches. In addition, a momentum update strategy is leveraged to update the weights of the key encoder (in $t$-th iteration) conditioned on query encoder weights: $w_{f_k}^{t} = \alpha \cdot w_{f_k}^{t-1} + (1 - \alpha)\cdot w_{f_q}^{t-1} $, where $w_{f_k}$ and $w_{f_q}$ are the weights of key encoder and query encoder. $\alpha$ is the momentum coefficient.

\section{Sequence Contrastive Learning}

\begin{figure*}[!tb]
\centering {\includegraphics[width=1\textwidth]{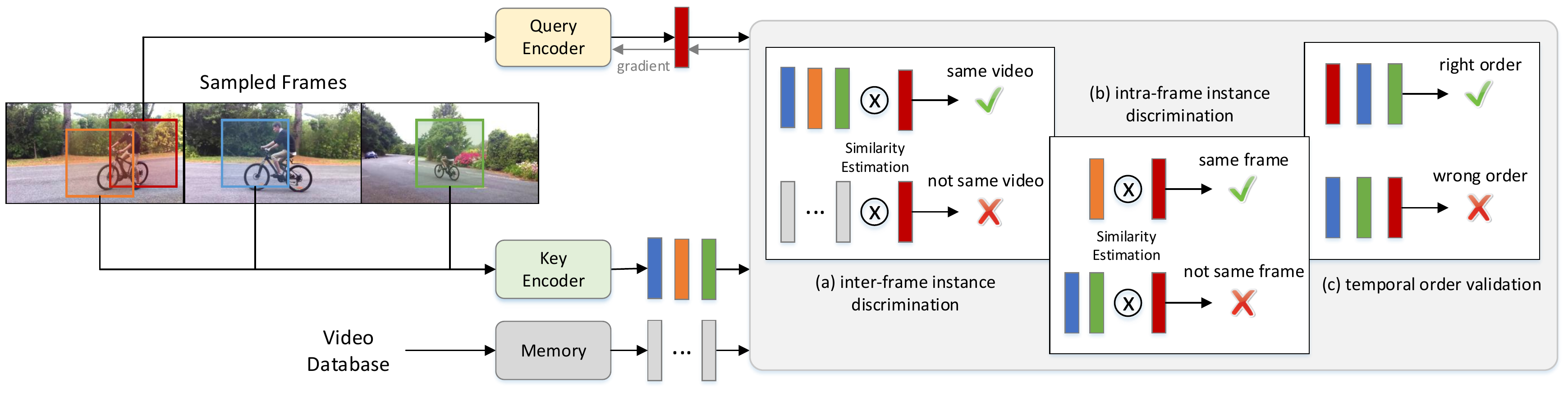}}
\caption{An overview of Sequence Contrastive Learning (SeCo) approach for unsupervised representation learning, which is composed of three proxy tasks: inter-frame instance discrimination task, intra-frame instance discrimination task, and temporal order validation task.}
\label{fig:figframework}
\end{figure*}

In this work, we remould the contrastive learning under the sequence supervision from videos, namely Sequence Contrastive Learning (SeCo), for unsupervised representation learning. In SeCo, three kinds of basic and generic supervision in the video sequence (from spatial, spatiotemporal, and sequential perspectives) are exploited to learn~powerful and generic visual representation. An overview of~our~sequence contrastive learning framework is illustrated in Figure \ref{fig:figframework}.

\subsection{Problem Formulation}

In the scenario of unsupervised video feature learning, we are given a collection of video sequences $\calV=\{v\}$ from a large-scale video benchmark. The goal is to pre-train a visual encoder over the video sequence data in an unsupervised manner to extract generic visual representations. The pre-trained visual encoder can be further utilized to support several video downstream~tasks.

Inspired by recent success of contrastive learning in image domain \cite{he2019momentum,wu2018unsupervised}, we formulate the unsupervised video feature learning in contrastive learning paradigm by exploiting the inherent supervision within sequential structure in videos. In particular, video is an information-intensive media with spatiotemporal coherence and variation across frames, which reflects three types of supervision from spatial, spatiotemporal and sequential perspectives. Accordingly, motivated by each type of supervision implicit in video sequence, we present one simple yet effective proxy task to guide the unsupervised feature learning with the corresponding supervision.

Formally, given an unlabeled video sequence $v$, we firstly sample three frames randomly ($s^1$, $s^2$, $s^3$) and take the first (or last) frame $s^1$ (or $s^3$) in time order as the anchor frame. The anchor frame is then transformed into two perturbed samples with different augmentations, one of which is taken as query $\bs_q$ and the other is used as key $\bs^1_k$. Meanwhile, we perform data augmentation over the other two frames, leading to two keys ($\bs^2_k$, $\bs^3_k$). In analogy to instance discrimination task in image domain that encourages a query matches a key if they are augmentations of an identical image, we consider \textbf{inter-frame instance discrimination task} that examines the compatibility of each query-key frame pair at video level, which is tailored for video understanding. That is, from the spatiotemporal perspective, the query $\bs_q$ should be similar to all the keys ($\bs^1_k$, $\bs^2_k$, $\bs^3_k$) in the same video, and dissimilar to the keys $\mathcal{K^-}$ sampled from other videos across mini-batches. Moreover, to characterize the temporal variation across frames in a video, a simple \textbf{intra-frame instance discrimination task} is particularly devised to determine whether two frame patches are derived from the same video frame, from the spatial standpoint. As such, the query $\bs_q$ is enforced to match key $\bs^1_k$ (augmented from the same frame $s^1$), and mismatch the keys ($\bs^2_k$, $\bs^3_k$) from other frames. Furthermore, from the sequential perspective, we involve the \textbf{temporal order validation task} to exploit the inherent sequential structure of videos by predicting the~correct temporal order of a frame patch sequence. Specifically, given the input frame patch sequence consisting of the query~$\bs_q$ and two keys ($\bs^2_k$, $\bs^3_k$), a linear classifier is leveraged to judge whether the query $\bs_q$ is in front of or behind keys ($\bs^2_k$, $\bs^3_k$).

\subsection{Inter-frame Instance Discrimination Task}
Unlike \cite{he2019momentum} that exploits image-level query-key compatibility, we facilitate contrastive learning in video domain via the inter-frame instance discrimination task, which aims to exploit the video-level query-key compatibility. In this proxy task, the pre-trained visual encoder is learnt to not only differentiate the two augmented frame patches of the same frame in a video from the negative/mismatched frame patches in other videos, but also recognize the patches of other frames in the same video as positive/matched samples. Such design goes beyond the traditional supervision in a static image with data augmentation, and fetches more positive frame patches within the same video as supervision for contrastive learning, which sheds new light on objects with temporal evolution (e.g., new views/poses of objects). The way elegantly takes the advantage of spatiotemporal structure within videos and thus strengthens the unsupervised visual feature learning for video understanding.

Technically, suppose we have the encoded query $\bs_q$ and key $\bs^1_k$ belonging to the same frame, and two keys ($\bs^2_k$, $\bs^3_k$) from other frames in the same video. In the inter-frame instance discrimination task, our target is to determine whether two frame patches are from the same video. Therefore, we define all the keys ($\bs^1_k$, $\bs^2_k$, $\bs^3_k$) within the same video as positive ones, and the frame patches sampled from other videos in neighboring mini-batches $\mathcal{K^-}$ are taken as the negative keys. Considering that the conventional formulation of contrastive learning (e.g., InfoNCE in Eq.(\ref{eq:cpc})) only penalizes the incompatibility of each positive query-key pair at a time, we derive a particular form of~contrastive learning that simultaneously match query $\bs_q$ to multiple positive keys ($\bs^1_k$, $\bs^2_k$, $\bs^3_k$) in our case.
In particular, the~new objective in this task is defined as the averaged sum of all the contrastive losses with regard to each positive~query-key pair ($\bs_q$,$\bs^i_k$):
\begin{equation}
\label{eq:inter}
\mathcal{L}_{Inter-frame}=\frac{1}{3}\sum_{i=1}^{3}{\mathcal{L}_{NCE}(\bs_q,\bs^i_k,\mathcal{K}^-)}.
\end{equation}
By minimizing the objective, the visual encoder is enforced to distinguish all the positive keys ($\bs^1_k$, $\bs^2_k$, $\bs^3_k$) and query $\bs_q$ within the same video from all the negative keys of other videos $\mathcal{K}^-$ at a time.

\subsection{Intra-frame Instance Discrimination Task}
In the inter-frame instance discrimination task, all sampled frame patches are holistically grouped as one generic class at video-level, while leaving the inherently spatial variation across frames within one video unexploited. To alleviate the issue, we additionally involve the intra-frame instance discrimination task to distinguish the frame patches of the same frame from the ones of the other frames in a video, which explicitly characters the variation from the spatial perspective. As such, by further steering unsupervised feature learning with the spatial supervision, the learnt visual representations are expected to be discriminative across frames in a video.

In particular, among all the four frame patches sampled from one video (query $\bs_q$ and key $\bs^1_k$ from an identical frame, and two keys $\bs^2_k$ \& $\bs^3_k$ from another two frames), we take $\bs^1_k$ as positive key and $\bs^2_k$ \& $\bs^3_k$ as negative keys with regard to query $\bs_q$. Note that since the previous proxy task has already exploited the incompatibility of negative query-key pairs derived from other videos, we exclude these negative keys for contrastive learning in this task for simplicity. Accordingly, we measure the objective of this task in the conventional form of contrastive loss:
\begin{equation}
\label{eq:intra}
\mathcal{L}_{Intra-frame}=\mathcal{L}_{NCE}(\bs_q,\bs^1_k,\{\bs^2_k,\bs^3_k\}).
\end{equation}
Such objective ensures that query $\bs_q$ is similar to the positive key $\bs^1_k$ augmented from the same frame and remains distinct to the negative keys $\{\bs^2_k,\bs^3_k\}$ of other frames, pursuing the temporally discriminative visual representation.

\subsection{Temporal Order Validation Task}
Most video applications (e.g., action recognition and object tracking) capitalize on the understanding of inherent sequential structure of videos, which can not be directly captured via the aforementioned two tasks that only exploit the spatiotemporal/spatial supervision based on individual frame patches. Therefore, we devise the temporal order validation task from a sequential perspective, aiming to verify whether a series of frame patches is in the correct temporal order. The rationale behind is to encourage the pre-trained visual encoder to reason about the temporal ordering of frame patches and thus exploit the sequential structure of videos for unsupervised feature learning.

Specifically, recall that we randomly sample three frames from an unlabeled video sequence and take the first or last frame in time order as the anchor frame, there are two kinds of temporal orders between query (augmented from anchor frame) and two keys (derived from the other two frames):\emph{ in front of} or \emph{behind}. Hence, given the input frame patch sequence consisting of query $\bs_q$ and two keys ($\bs^2_k$, $\bs^3_k$), we concatenate the query and two keys as the holistic sequence representation and feed it into a binary classifier $g(\cdot)$, which predicts if the query is in front of or behind keys. The whole model is thus optimized with cross-entropy loss:
\begin{equation}\small
\label{eq:order}
\mathcal{L}_{Temporal} = - y \log g(\bs_q,\bs^2_k,\bs^3_k) - (1-y) \log (1-g(\bs_q,\bs^2_k,\bs^3_k)),
\end{equation}
where $y\in \{0,1\}$ represents the ground-truth label that indicates whether the query $\bs_q$ is in front of or behind the two~keys ($\bs^2_k$, $,\bs^3_k$).

\subsection{Optimization}
\textbf{Training Objective.}
The overall training objective of our sequence contrastive learning integrates all the objectives of three proxy tasks (i.e., Eq.(\ref{eq:inter}) for inter-frame instance discrimination task, Eq.(\ref{eq:intra}) for intra-frame instance~discrimination task, and Eq.(\ref{eq:order}) for temporal order validation task):
\begin{equation}\small
\label{eq:overall}
\mathcal{L} = \mathcal{L}_{Inter-frame}+\mathcal{L}_{Intra-frame}+\mathcal{L}_{Temporal}.
\end{equation}

\textbf{Weights Update.}
In our SeCo, the query encoder $f_q$ is directly optimized with standard SGD algorithm by minimizing $\mathcal{L}$. The weights of key encoder $f_k$ is accordingly updated conditioned on query encoder weights via a momentum update strategy:
\begin{equation}\small
\label{eq:update}
w_{f_k}^{t} = \alpha \cdot w_{f_k}^{t-1} + (1 - \alpha)\cdot w_{f_q}^{t-1},
\end{equation}
where $\alpha$ denotes the momentum coefficient that controls the updating of key encoder weights.

\section{Experiments}
We empirically verify the merit of SeCo for unsupervised representation learning in three downstream tasks: action recognition, untrimmed activity recognition and object tracking. The first experiment is conducted respectively on action recognition (Kinetics), untrimmed activity recognition (ActivityNet) and object tracking (OTB-100) under ``pre-trained representation + linear model'' protocol. The second experiment transfers the network unsupervised pre-trained by SeCo as the initialization for fine-tuning in action recognition task (UCF101 and HMDB51). That is ``pre-training + fine-tuning''~protocol.

\subsection{Datasets}
\textbf{Kinetics400} dataset \cite{kay2017the} is one of the large-scale action recognition benchmarks which contains around 300K videos from 400 action categories. Each video clip in this dataset is cropped from the raw YouTube video and the duration is 10 seconds. All the videos are grouped into three subsets for training (240K), validation (20K), and testing (40K), respectively. Because the labels of testing set are not publicly available, the performances on the Kinetics400 dataset are reported on the validation set. \textbf{UCF101} \cite{soomro2012ucf101} is one of the most popular action recognition benchmarks. This dataset consists of 13,320 videos from 101 action classes, which are split into about 9.5K and 3.7K videos in training and testing set, respectively. \textbf{HMDB51} \cite{kuehne2011hmdb} is another widely used action recognition dataset and includes 7K videos from 51 action~categories. The dataset is split into training (3.5K) and testing (1.5K) sets.

\textbf{ActivityNet} dataset \cite{heilbron2015activitynet} is a large-scale human activity understanding benchmark. The latest released version (v1.3) consists of 19,994 videos from 200 activity categories and is utilized here for evaluation. All the videos in the dataset are divided into 10,024, 4,926, and 5,044 for training, validation, and testing sets, respectively.~The labels of testing set are not publicly available and~thus the performances on ActivityNet dataset are all reported on validation set.

The task of object tracking actually involves two widely adopted datasets in our case, including Generic Object Tracking Benchmark (GOT-10K \cite{huang2019got}) and Object Tracking Benchmark 2015 (OTB-100 \cite{wu2015object}). \textbf{GOT-10K} dataset contains more than 10K real-world videos with moving objects and over 1.5M manually labeled bounding boxes. The dataset covers more than 560 categories of moving objects and 80+ categories of motion patterns. We exploit the training set of 9,335 videos to learn a linear feature transformer ($1\times1$ convolution), whose outputs serve for the template matching in feature space to track the example object in the subsequent~frames. \textbf{OTB-100} dataset includes 100 video sequences, which are utilized as the test set for the evaluation of object tracking.

\begin{figure*}[!tb]
\centering {\includegraphics[width=0.86\textwidth]{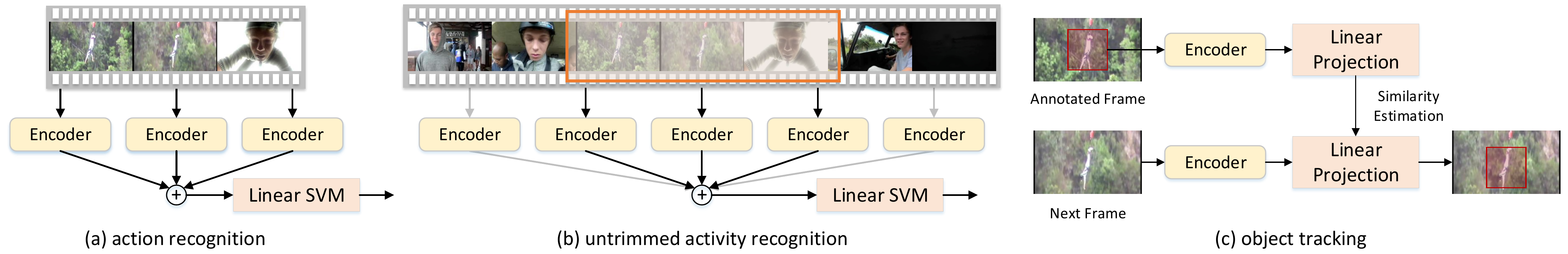}}
\caption{The detailed procedures of three downstream tasks, i.e., (a) action recognition, (b) untrimmed activity recognition, and (c) object tracking, under ``pre-trained representation + linear model'' protocol.}
\label{fig:task}
\end{figure*}

\begin{table*}[]
  \caption{Performance comparison of the representations pre-trained by different mechanisms in three downstream tasks under ``Pre-trained Representation + Linear Model'' protocol.}
  \label{sample-table}
  \setlength\tabcolsep{12pt}
  \centering
  \begin{tabular}{lcccc}\toprule
                           & Action       & Untrimmed Activity & \multicolumn{2}{c}{Object}          \\
                           & Recognition  & Recognition        & \multicolumn{2}{c}{Tracking}        \\\midrule
    Dataset                & Kinetics 400 & ActivityNet        & \multicolumn{2}{c}{OTB-100}         \\
    Learnable Module       & Linear SVM   & Linear SVM         & \multicolumn{2}{c}{1x1 Convolution} \\
    Metric                 & Top-1        & Top-1              & Precision          & Success                      \\\midrule
    MoCo-ImageNet          & 51.30        & 66.17              & 59.91              & 43.06                       \\
    Supervised ImageNet Pre-training    & 52.34        & 67.19 & 69.54              & 48.01                       \\
    VINCE~\cite{gordon2020watching}     & 36.20        & -     & 62.90              & 46.50                       \\\midrule
    SeCo-Inter             & 58.97        & 66.69              & 67.92              & 48.03                       \\
    SeCo-Inter+Intra       & 60.74        & 68.31              & 70.29              & 50.48                       \\
    SeCo-Inter+Intra+Order & \textbf{61.91}        & \textbf{68.55}            & \textbf{71.86}              & \textbf{51.78}                       \\
    \bottomrule
  \end{tabular}
\end{table*}

\subsection{Experimental Settings}
\textbf{SeCo Training.} We perform the training of our SeCo on the training set of Kinetics400 dataset and utilize the backbone of ResNet50 plus an MLP head. Note that the MLP head only influences SeCo training and is not involved in downstream tasks. The image patches input to the backbone are with the size of $224\times 224$, and the head takes the global pooling feature as the input and embeds the feature into 128$d$ with two fully-connected layers ($2048\times 2048$ and $2048\times 128$). The output vector of the MLP head is normalized by its L2-norm and then exploited as the encoded representation of query or keys. In our implementations, the size of the mini-batch is set to 512 and the size of memory is 131,072. The momentum coefficient $\alpha$ for momentum update of the encoder is set to 0.999 and the temperature $\tau$ in infoNCE loss is 0.1. Following~\cite{he2019momentum}, shuffling BN is utilized for multi-GPU training. To optimize the parameters in the encoder, we use the momentum SGD with~initial learning rate 0.2 which is annealed down to zero following a cosine decay. The network is trained for 400 epoch base on the network initialized with MoCo~\cite{he2019momentum} on ImageNet. For data augmentation, we employ random cropping with random scales, color-jitter, random grayscale, blur, and mirror.

\textbf{Action Recognition and Untrimmed Activity Recognition under ``Pre-trained Representation + Linear Model'' Protocol.} We directly exploit the backbone of unsupervised learnt network by SeCo on Kinetics400 as the feature extractor, and verify the frozen representation via linear classification on both downstream tasks of action recognition and untrimmed activity recognition. For each video in Kinetics400 and ActivityNet, we uniformly sample 30 and 50 frames, respectively, resize each frame with short edge of 256, and crop the resized version to $224\times 224$ by using center crop. As shown in Figure \ref{fig:task} (a) and (b), we extract the~frame-level feature by feature extractor and average all the frame-level features to obtain the video-level representation. A linear SVM is finally trained on the training videos of Kinetics400 or ActivityNet and evaluated on each validation set. We adopt the top-1 accuracy as the performance metric of the two tasks.

\textbf{Object Tracking under ``Pre-trained Representation + Linear Model'' Protocol.} Given the initial bounding box of an object in the first frame of a video, the task of object tracking is to locate the object in the subsequent frames. We exploit SiamFC~\cite{gordon2020watching} as our tracking algorithm and execute object tracking on the representation pre-learnt by SeCo, as illustrated in Figure \ref{fig:task} (c). Following the setting in SiamFC that the spatial resolution of the output feature map is 1/8 of the input image, we modify the configuration of ResNet50. Specifically, for the convolution with ``stride 2'' in the last two stages $\{res4,res5\}$, the ``stride'' is changed to 1, and for the $3\times 3$ convolutions in $res4$ and $res5$, the dilation rate is modified from 1 to 2 and 4, respectively. Note that the weights of the layers in ResNet50 remain unchanged during such modification and thus the representations are still considered as frozen. Furthermore, an additional $1\times 1$ convolution is placed on the top of the backbone to transform the frozen representation for SiamFC tracking algorithm. In this sense, only $1\times 1$ convolution is learnable and we also regard such protocol as linear model. The $1\times 1$ convolution is optimized with the training set of GOT-10K, and object tracking is evaluated on OTB-100 in terms of two performance metrics: Area Under the Curve (AUC) of precision and success.

\textbf{Action Recognition with ``Pre-training + Fine-tuning'' Protocol.} Another essential function of unsupervised learning is for the purpose of network pre-training, which serves as the network initialization for fine-tuning in downstream tasks. Here, we initialize ResNet50 with the backbone in the unsupervised training of SeCo and fine-tune the network with the standard supervised setting \cite{qiu2017learning,qiu2019learning} on UCF101 and HMDB51 for action recognition.

\begin{table}[]
  \caption{Performance comparisons of unsupervised representation learning on Kinetics400.}
  \label{tab.kinetics_unsup}
  \centering
 \begin{tabular}{l@{~~}c}\toprule
    Method            & Top-1 (\%) \\\midrule
    OPN$^{\dagger}$~\cite{lee2017unsupervised}        & 20.86         \\
    RotNet$^{\dagger}$~\cite{gidaris2018unsupervised} & 23.33         \\
    3DRotNet$^{\dagger}$~\cite{Jing2018SelfSupervisedSF}          & 19.33         \\
    VIE-Single~\cite{zhuang2019unsupervised}          & 44.41         \\
    VIE-TRN~\cite{zhuang2019unsupervised}             & 44.91         \\
    VIE-3DResNet~\cite{zhuang2019unsupervised}        & 43.40         \\
    VIE-SlowFast~\cite{zhuang2019unsupervised}        & 47.37         \\
    VIE-Full~\cite{zhuang2019unsupervised}            & 48.53         \\\midrule
    SeCo (ResNet18)                                   & \textbf{50.81}         \\
    \bottomrule
  \end{tabular}
\end{table}

\subsection{Evaluations on Pre-trained Representation + Linear Model protocol}
We first validate our SeCo under the protocol of ``Pre-trained Representation + Linear Model,'' which is to manifest the generalization capability of representations learnt by SeCo. We compare the following three training mechanisms: (1) MoCo-ImageNet train the network on ImageNet in an unsupervised manner by using MoCo~\cite{he2019momentum} algorithm. (2) Supervised ImageNet Pre-training capitalizes on the supervision of human-annotated labels on the images and learns the network in a fully-supervised fashion. (3) VINCE~\cite{gordon2020watching} forms multiple anchor-positive pairs from multiple frames in a video and also executes contrastive training for unsupervised representation learning.

Table \ref{sample-table} summarizes performance comparisons of different representation learning mechanisms in three downstream tasks. Overall, the performances across the three tasks consistently indicate that our SeCo leads to performance boost against other training mechanisms. Particularly, by doing classification on the representations pre-learnt by SeCo achieves 61.91\% and 68.55\% on action recognition (Kinetics400) and untrimmed activity recognition (ActivityNet), respectively, making the absolute improvement over Supervised ImageNet Pre-training by 9.57\% and 1.36\% in terms of top-1 accuracy. Furthermore, SeCo benefits from three types of supervision, and models the spatiotemporal coherence and variation in videos better, therefore leading the precision by 2.32\% in object tracking (OTB-100). The results clearly demonstrate the advantage of our SeCo unsupervised pre-training for learning representations that are more generic across various downstream tasks. As expected, SeCo-Inter remoulds MoCo-ImageNet in the context of video and exhibits better performance than MoCo-ImageNet on video tasks. SeCo-Inter+Intra constantly outperforms SeCo-Inter and SeCo learnt through the three proxy tasks performs the best. The results also verify the complementarity between three supervision in the sequence for representation~learning.

Table \ref{tab.kinetics_unsup} further details the comparisons with state-of-the-art unsupervised representation learning methods on Kinetics400. $^{\dagger}$ denotes that each method is implemented and learnt on Kinetics400 as reported in \cite{zhuang2019unsupervised}. Please also note that here we exploit ResNet18 as the backbone in our SeCo training for fair comparisons. Specifically, VIE learns deep nonlinear embeddings to group similar videos and push different videos apart in the embedding space and such idea is similar to our SeCo-Inter in spirit. As indicated by the results, VIE-Single leads to a large performance gain over OPN and RotNet, and all the three runs select one frame from each video, which is input into a 2D network for classification. VIE-3DResNet further extends 2D ResNet18 to 3D and VIE-SlowFast employs the advanced SlowFast structure of two 3D networks. By combining VIE-Single and VIE-SlowFast, VIE-Full achieves 48.53\% top-1 accuracy, which is still lower than 50.81\% of SeCo learnt only on a 2D ResNet18. That again proves the impact of our SeCo for unsupervised representation learning.

\subsection{Evaluations on Pre-training + Fine-tuning Protocol}
Next, we evaluate SeCo from the aspect of network pre-training, which is taken as network initialization for fine-tuning on downstream tasks. Such protocol is to examine the transferability of the pre-trained structure. Table \ref{tab.ucf_hmdb_tune} shows the comparisons of pre-training the networks by different methods and then supervised fine-tuning on UCF101 and HMDB51 as the backbone in TSN \cite{wang2018temporal} for action recognition. Compared to the best competitor VIE-Full, SeCo improves the top-1 accuracy from 80.40\%/52.50\% to 88.26\%/55.55\% on the two datasets. Notably, SeCo unsupervised pre-training leads the accuracy by 2.96\% and 6.47\% against fully-supervised ImageNet pre-training, which is really impressive.

\begin{table}[]
  \caption{Performance comparisons of pre-training + fine-tuning protocol on UCF101 and HMDB51.}
  \label{tab.ucf_hmdb_tune}
  \centering
 \begin{tabular}{l@{~~}c@{~~}c}\toprule
                              & \multicolumn{2}{c}{ Top-1 (\%)} \\
                              & UCF101         & HMDB51         \\\midrule
    Shuffle\&Learn~\cite{misra2016shuffle}     & 50.20            & 18.10          \\
    OPN~\cite{lee2017unsupervised}             & 59.60            & 23.80          \\
    ClipOrder~\cite{xu2019self}                & 72.40            & 30.90          \\
    3DRotNet~\cite{Jing2018SelfSupervisedSF}               & 66.00            & 37.10          \\
    DPC~\cite{Han2019VideoRL}                  & 75.70            & 35.70          \\
    CBT~\cite{sun2019learning}                 & 79.50            & 44.60          \\
    VIE-Full~\cite{zhuang2019unsupervised}     & 80.40            & 52.50          \\ \midrule
    Supervised ImageNet Pre-training           & 85.30 & 49.08\\
    SeCo                      & \textbf{88.26}              & \textbf{55.55}            \\
    \bottomrule
    \end{tabular}
\end{table}

\section{Conclusions}
We have presented Sequence Contrastive Learning (SeCo) method which explores the generic supervision in the video sequence for unsupervised representation learning. Particularly, we study the sequence supervision systematically from three aspects: spatial, spatiotemporal and sequential. To verify our claim, we devise one simple proxy task, i.e., intra-frame/inter-frame instance discrimination task or temporal order validation task, to present and leverage each supervision. In between, intra-frame/inter-frame instance discrimination task is to determine whether two frame patches are from one frame or an identical video, respectively, and temporal order validation examines whether a series of frame patches are in chronological order correctly. We materialize the three proxy tasks and build our SeCo on contrastive learning framework. Experiments conducted on both ``pre-trained representation + linear model'' and ``pre-training + fine-tuning'' protocols, validate our proposal and analysis. More remarkably, SeCo pre-training leads to an increase of accuracy by 2.96\% and 6.47\% over ImageNet supervised~pre-training on UCF101 and HMDB51 datasets for action recognition task.

\section{Ethical Statement}
Video understanding (e.g., action recognition and object tracking) is one of the fundamental problems in numerous real-world applications, ranging from video surveillance, indexing and retrieval to human computer interaction. However, the achievements of these video applications rely heavily on the assumption that large quantities of human annotations are accessible for neural model learning. The assumption becomes impractical when cost-expensive and labor-intensive manual labeling is required. This significantly limits and discourages the motivations for relatively small research communities without adequate financial supports. We demonstrate in this paper that the challenge can be mitigated by pre-training a visual encoder via our Sequence Contrastive Learning (SeCo) in an unsupervised manner without any human-annotated labels. Such pre-trained visual encoder can be further utilized to facilitate a wide variety of video applications. Notice that our SeCo, an unsupervised learning approach, even surpasses the supervised ImageNet pre-training counterpart in action recognition task. Nevertheless, one potential risk lies in that if the use of unsupervised visual representation learning in videos means video understanding systems may now be easily developed by those with lower levels of domain or ML expertise, this could increase the risk of the video understanding model or its outputs being used incorrectly.

\bibliographystyle{aaai21}
\bibliography{egbib}

\begin{thebibliography}{44}
\providecommand{\natexlab}[1]{#1}
\providecommand{\url}[1]{\texttt{#1}}
\providecommand{\urlprefix}{URL }
\expandafter\ifx\csname urlstyle\endcsname\relax
  \providecommand{\doi}[1]{doi:\discretionary{}{}{}#1}\else
  \providecommand{\doi}{doi:\discretionary{}{}{}\begingroup
  \urlstyle{rm}\Url}\fi

\bibitem[{{Agrawal}, {Carreira}, and {Malik}(2015)}]{agrawal2015learning}
{Agrawal}, P.; {Carreira}, J.; and {Malik}, J. 2015.
\newblock Learning to See by Moving.
\newblock In \emph{ICCV}.

\bibitem[{Ahsan, Madhok, and Essa(2019)}]{Ahsan2019VideoJU}
Ahsan, U.; Madhok, R.; and Essa, I. 2019.
\newblock Video Jigsaw: Unsupervised Learning of Spatiotemporal Context for
  Video Action Recognition.
\newblock In \emph{WACV}.

\bibitem[{{Bachman}, {Hjelm}, and {Buchwalter}(2019)}]{bachman2019learning}
{Bachman}, P.; {Hjelm}, R.~D.; and {Buchwalter}, W. 2019.
\newblock Learning Representations by Maximizing Mutual Information Across
  Views.
\newblock In \emph{NeurIPS}.

\bibitem[{Cai et~al.(2020)Cai, Wang, Pan, Yao, and Mei}]{cai2020joint}
Cai, Q.; Wang, Y.; Pan, Y.; Yao, T.; and Mei, T. 2020.
\newblock Joint Contrastive Learning with Infinite Possibilities.
\newblock In \emph{NeurIPS}.

\bibitem[{{Deshpande}, {Rock}, and {Forsyth}(2015)}]{deshpande2015learning}
{Deshpande}, A.; {Rock}, J.; and {Forsyth}, D. 2015.
\newblock Learning Large-Scale Automatic Image Colorization.
\newblock In \emph{ICCV}.

\bibitem[{{Doersch}, {Gupta}, and {Efros}(2015)}]{doersch2015unsupervised}
{Doersch}, C.; {Gupta}, A.; and {Efros}, A.~A. 2015.
\newblock Unsupervised Visual Representation Learning by Context Prediction.
\newblock In \emph{ICCV}.

\bibitem[{{Finn}, {Goodfellow}, and {Levine}(2016)}]{finn2016unsupervised}
{Finn}, C.; {Goodfellow}, I.~J.; and {Levine}, S. 2016.
\newblock Unsupervised Learning for Physical Interaction through Video
  Prediction.
\newblock In \emph{NIPS}.

\bibitem[{{Gidaris}, {Singh} et~al.(2018)}]{gidaris2018unsupervised}
{Gidaris}, S.; {Singh}, P.; et~al. 2018.
\newblock Unsupervised Representation Learning by Predicting Image Rotations.
\newblock In \emph{ICLR}.

\bibitem[{{Gordon} et~al.(2020){Gordon}, {Ehsani}, {Fox}, and
  {Farhadi}}]{gordon2020watching}
{Gordon}, D.; {Ehsani}, K.; {Fox}, D.; and {Farhadi}, A. 2020.
\newblock Watching the World Go By: Representation Learning from Unlabeled
  Videos.
\newblock \emph{arXiv preprint arXiv:2003.07990} .

\bibitem[{{Goyal} et~al.(2019){Goyal}, {Mahajan}, {Gupta}, and
  {Misra}}]{goyal2019scaling}
{Goyal}, P.; {Mahajan}, D.; {Gupta}, A.; and {Misra}, I. 2019.
\newblock Scaling and Benchmarking Self-Supervised Visual Representation
  Learning.
\newblock In \emph{ICCV}.

\bibitem[{Han, Xie et~al.(2019)}]{Han2019VideoRL}
Han, T.; Xie, W.; et~al. 2019.
\newblock Video Representation Learning by Dense Predictive Coding.
\newblock In \emph{ICCV Workshop}.

\bibitem[{{He} et~al.(2019){He}, {Fan}, {Wu}, {Xie}, and
  {Girshick}}]{he2019momentum}
{He}, K.; {Fan}, H.; {Wu}, Y.; {Xie}, S.; and {Girshick}, R.~B. 2019.
\newblock Momentum Contrast for Unsupervised Visual Representation Learning.
\newblock \emph{arXiv preprint arXiv:1911.05722} .

\bibitem[{{Heilbron} et~al.(2015){Heilbron}, {Escorcia}, {Ghanem}, and
  {Niebles}}]{heilbron2015activitynet}
{Heilbron}, F.~C.; {Escorcia}, V.; {Ghanem}, B.; and {Niebles}, J.~C. 2015.
\newblock ActivityNet: A large-scale video benchmark for human activity
  understanding.
\newblock In \emph{CVPR}.

\bibitem[{{Hjelm} et~al.(2019){Hjelm}, {Fedorov}, {Lavoie-Marchildon},
  {Grewal}, {Bachman}, {Trischler}, and {Bengio}}]{hjelm2019learning}
{Hjelm}, R.~D.; {Fedorov}, A.; {Lavoie-Marchildon}, S.; {Grewal}, K.;
  {Bachman}, P.; {Trischler}, A.; and {Bengio}, Y. 2019.
\newblock Learning deep representations by mutual information estimation and
  maximization.
\newblock In \emph{ICLR}.

\bibitem[{{Huang}, {Zhao}, and {Huang}(2019)}]{huang2019got}
{Huang}, L.; {Zhao}, X.; and {Huang}, K. 2019.
\newblock GOT-10k: A Large High-Diversity Benchmark for Generic Object Tracking
  in the Wild.
\newblock \emph{IEEE Trans. on PAMI} .

\bibitem[{Jayaraman and Grauman(2015)}]{jayaraman2015learning}
Jayaraman, D.; and Grauman, K. 2015.
\newblock Learning image representations tied to ego-motion.
\newblock In \emph{ICCV}.

\bibitem[{Jing et~al.(2018)Jing, Yang, Liu, and
  Tian}]{Jing2018SelfSupervisedSF}
Jing, L.; Yang, X.; Liu, J.; and Tian, Y. 2018.
\newblock Self-Supervised Spatiotemporal Feature Learning via Video Rotation
  Prediction.
\newblock \emph{arXiv preprint arXiv:1811.11387} .

\bibitem[{{Kay} et~al.(2017){Kay}, {Carreira}, {Simonyan}, {Zhang}, {Hillier},
  {Vijayanarasimhan}, {Viola}, {Green}, {Back}, {Natsev}, {Suleyman}, and
  {Zisserman}}]{kay2017the}
{Kay}, W.; {Carreira}, J.; {Simonyan}, K.; {Zhang}, B.; {Hillier}, C.;
  {Vijayanarasimhan}, S.; {Viola}, F.; {Green}, T.; {Back}, T.; {Natsev}, P.;
  {Suleyman}, M.; and {Zisserman}, A. 2017.
\newblock The Kinetics Human Action Video Dataset.
\newblock \emph{arXiv preprint arXiv:1705.06950} .

\bibitem[{Kuehne et~al.(2011)Kuehne, Jhuang, Garrote, Poggio, and
  Serre}]{kuehne2011hmdb}
Kuehne, H.; Jhuang, H.; Garrote, E.; Poggio, T.; and Serre, T. 2011.
\newblock HMDB: A large video database for human motion recognition.
\newblock In \emph{ICCV}.

\bibitem[{{Lee} et~al.(2017){Lee}, {Huang}, {Singh}, and
  {Yang}}]{lee2017unsupervised}
{Lee}, H.-Y.; {Huang}, J.-B.; {Singh}, M.; and {Yang}, M.-H. 2017.
\newblock Unsupervised Representation Learning by Sorting Sequences.
\newblock In \emph{ICCV}.

\bibitem[{Liu et~al.(2017)Liu, Yeh, Tang, Liu, and Agarwala}]{Liu2017VideoFS}
Liu, Z.; Yeh, R.~A.; Tang, X.; Liu, Y.; and Agarwala, A. 2017.
\newblock Video Frame Synthesis Using Deep Voxel Flow.
\newblock In \emph{ICCV}.

\bibitem[{{Lorre} et~al.(2020){Lorre}, {Rabarisoa}, {Orcesi}, {Ainouz}, and
  {Canu}}]{lorre2020temporal}
{Lorre}, G.; {Rabarisoa}, J.; {Orcesi}, A.; {Ainouz}, S.; and {Canu}, S. 2020.
\newblock Temporal Contrastive Pretraining for Video Action Recognition.
\newblock In \emph{WACV}.

\bibitem[{Luo et~al.(2017)Luo, Peng, Huang, Alahi, and
  Fei-Fei}]{Luo2017UnsupervisedLO}
Luo, Z.; Peng, B.; Huang, D.-A.; Alahi, A.; and Fei-Fei, L. 2017.
\newblock Unsupervised Learning of Long-Term Motion Dynamics for Videos.
\newblock In \emph{CVPR}.

\bibitem[{{Misra} et~al.(2016)}]{misra2016shuffle}
{Misra}, I.; et~al. 2016.
\newblock Shuffle and Learn: Unsupervised Learning Using Temporal Order
  Verification.
\newblock In \emph{ECCV}.

\bibitem[{{Mobahi}, {Collobert}, and {Weston}(2009)}]{mobahi2009deep}
{Mobahi}, H.; {Collobert}, R.; and {Weston}, J. 2009.
\newblock Deep learning from temporal coherence in video.
\newblock In \emph{ICML}.

\bibitem[{{Noroozi} and {Favaro}(2016)}]{noroozi2016unsupervised}
{Noroozi}, M.; and {Favaro}, P. 2016.
\newblock Unsupervised Learning of Visual Representations by Solving Jigsaw
  Puzzles.
\newblock In \emph{ECCV}.

\bibitem[{Oord, Li, and Vinyals(2018)}]{oord2018representation}
Oord, A. v.~d.; Li, Y.; and Vinyals, O. 2018.
\newblock Representation learning with contrastive predictive coding.
\newblock \emph{arXiv preprint arXiv:1807.03748} .

\bibitem[{Pan et~al.(2016)Pan, Li, Yao, Mei, Li, and Rui}]{pan2016learning}
Pan, Y.; Li, Y.; Yao, T.; Mei, T.; Li, H.; and Rui, Y. 2016.
\newblock Learning deep intrinsic video representation by exploring temporal
  coherence and graph structure.
\newblock In \emph{IJCAI}.

\bibitem[{Qiu, Yao, and Mei(2017)}]{qiu2017learning}
Qiu, Z.; Yao, T.; and Mei, T. 2017.
\newblock Learning Spatio-Temporal Representation with Pseudo-3D Residual
  Networks.
\newblock In \emph{ICCV}.

\bibitem[{Qiu et~al.(2019)Qiu, Yao, Ngo, Tian, and Mei}]{qiu2019learning}
Qiu, Z.; Yao, T.; Ngo, C.-W.; Tian, X.; and Mei, T. 2019.
\newblock Learning spatio-temporal representation with local and global
  diffusion.
\newblock In \emph{CVPR}.

\bibitem[{{Soomro}, {Zamir}, and {Shah}(2012)}]{soomro2012ucf101}
{Soomro}, K.; {Zamir}, A.~R.; and {Shah}, M. 2012.
\newblock UCF101: A Dataset of 101 Human Actions Classes From Videos in The
  Wild.
\newblock \emph{arXiv preprint arXiv:1212.0402} .

\bibitem[{Srivastava, Mansimov, and
  Salakhutdinov(2015)}]{Srivastava2015UnsupervisedLO}
Srivastava, N.; Mansimov, E.; and Salakhutdinov, R. 2015.
\newblock Unsupervised Learning of Video Representations using LSTMs.
\newblock In \emph{ICML}.

\bibitem[{{Sun} et~al.(2019){Sun}, {Baradel}, {Murphy}, and
  {Schmid}}]{sun2019learning}
{Sun}, C.; {Baradel}, F.; {Murphy}, K.; and {Schmid}, C. 2019.
\newblock Learning Video Representations using Contrastive Bidirectional
  Transformer.
\newblock \emph{arXiv preprint arXiv:1906.05743} .

\bibitem[{{Tian}, {Krishnan}, and {Isola}(2019)}]{tian2019contrastive}
{Tian}, Y.; {Krishnan}, D.; and {Isola}, P. 2019.
\newblock Contrastive Multiview Coding.
\newblock \emph{arXiv preprint arXiv:1906.05849} .

\bibitem[{Vondrick, Pirsiavash, and Torralba(2016)}]{Vondrick2016GeneratingVW}
Vondrick, C.; Pirsiavash, H.; and Torralba, A. 2016.
\newblock Generating Videos with Scene Dynamics.
\newblock In \emph{NIPS}.

\bibitem[{Wang et~al.(2019)Wang, Jiao, Bao, He, Liu, and
  Liu}]{Wang2019SelfSupervisedSR}
Wang, J.; Jiao, J.; Bao, L.; He, S.; Liu, Y.; and Liu, W. 2019.
\newblock Self-Supervised Spatio-Temporal Representation Learning for Videos by
  Predicting Motion and Appearance Statistics.
\newblock In \emph{CVPR}.

\bibitem[{Wang et~al.(2018)Wang, Xiong, Wang, Qiao, Lin, Tang, and
  Van~Gool}]{wang2018temporal}
Wang, L.; Xiong, Y.; Wang, Z.; Qiao, Y.; Lin, D.; Tang, X.; and Van~Gool, L.
  2018.
\newblock Temporal segment networks for action recognition in videos.
\newblock \emph{IEEE Trans. on PAMI} .

\bibitem[{{Wang} and {Gupta}(2015)}]{wang2015unsupervised}
{Wang}, X.; and {Gupta}, A. 2015.
\newblock Unsupervised Learning of Visual Representations Using Videos.
\newblock In \emph{ICCV}.

\bibitem[{{Wu}, {Lim}, and {Yang}(2015)}]{wu2015object}
{Wu}, Y.; {Lim}, J.; and {Yang}, M.-H. 2015.
\newblock Object Tracking Benchmark.
\newblock \emph{IEEE Trans. on PAMI} .

\bibitem[{{Wu} et~al.(2018){Wu}, {Xiong}, {Yu}, and {Lin}}]{wu2018unsupervised}
{Wu}, Z.; {Xiong}, Y.; {Yu}, S.~X.; and {Lin}, D. 2018.
\newblock Unsupervised Feature Learning via Non-parametric Instance
  Discrimination.
\newblock In \emph{CVPR}.

\bibitem[{{Xu} et~al.(2019){Xu}, {Xiao}, {Zhao}, {Shao}, {Xie}, and
  {Zhuang}}]{xu2019self}
{Xu}, D.; {Xiao}, J.; {Zhao}, Z.; {Shao}, J.; {Xie}, D.; and {Zhuang}, Y. 2019.
\newblock Self-Supervised Spatiotemporal Learning via Video Clip Order
  Prediction.
\newblock In \emph{CVPR}.

\bibitem[{{Zhang}, {Isola}, and {Efros}(2016)}]{zhang2016colorful}
{Zhang}, R.; {Isola}, P.; and {Efros}, A.~A. 2016.
\newblock Colorful Image Colorization.
\newblock In \emph{ECCV}.

\bibitem[{{Zhuang} et~al.(2019){Zhuang}, {She}, {Andonian}, and
  {Yamins}}]{zhuang2019unsupervised}
{Zhuang}, C.; {She}, T.; {Andonian}, A.; and {Yamins}, D. 2019.
\newblock Unsupervised Learning from Video with Deep Neural Embeddings.
\newblock \emph{arXiv preprint arXiv:1905.11954} .

\bibitem[{{Zou} et~al.(2012){Zou}, {Zhu}, {Yu}, and {Ng}}]{zou2012deep}
{Zou}, W.; {Zhu}, S.; {Yu}, K.; and {Ng}, A.~Y. 2012.
\newblock Deep Learning of Invariant Features via Simulated Fixations in Video.
\newblock In \emph{NIPS}.

\end{thebibliography}

\end{document}